\documentclass{article}
\usepackage{spconf,amsmath,graphicx}
\usepackage{multirow}
\usepackage{booktabs}
\usepackage{amssymb}
\usepackage{amsmath}
\usepackage{xspace}
\usepackage{color}
\usepackage{pifont}
\usepackage{hyperref}

\newcommand{\methodname}{CHS-Net\xspace}

\newcommand{\mat}[1]{\boldsymbol{#1}}
\newcommand{\ie}{\emph{i.e.}\xspace}
\newcommand{\etal}{\emph{et al.}\xspace}

\newcommand{\cmark}{\ding{51}}%
\newcommand{\xmark}{\ding{55}}%

\newlength\savewidth\newcommand\shline{\noalign{\global\savewidth\arrayrulewidth
  \global\arrayrulewidth 1.5pt}\hline\noalign{\global\arrayrulewidth\savewidth}}
\title{cross-head supervision for crowd counting with noisy annotations}
\name{Mingliang Dai$^{1}$ \qquad Zhizhong Huang$^{1}$ \qquad Jiaqi Gao$^{1}$ \qquad Hongming Shan$^{2}$ \qquad Junping Zhang$^{1\dagger}$\thanks{$\dagger$: Corresponding author.  jpzhang@fudan.edu.cn. This work was supported in part by the National Natural Science Foundation of China (Nos. 62176059 and 62101136), the Shanghai Municipal Science and Technology
Major Project (No. 2018SHZDZX01), the Zhangjiang Laboratory (ZJLab) and the Shanghai Center for Brain Science and Brain-Inspired Technology.}}
\address{$^1$ Shanghai Key Lab of Intelligent Information Processing, School of Computer Science\\
$^2$ Institute of Science and Technology for Brain-inspired Intelligence\\
Fudan University, Shanghai 200433, China}
\begin{document}
%
\maketitle
\begin{abstract}
Noisy annotations such as missing annotations and location shifts often exist in crowd counting datasets due to multi-scale head sizes, high occlusion, etc.
These noisy annotations severely affect the model training, especially for density map-based methods. 
To alleviate the negative impact of noisy annotations, we propose a novel crowd counting model with one convolution head and one transformer head, in which these two heads can supervise each other in noisy areas, called \textbf{C}ross-\textbf{H}ead \textbf{S}upervision. The resultant model, \methodname, can synergize different types of inductive biases for better counting.
In addition, we develop a progressive cross-head supervision learning strategy to stabilize the training process and provide more reliable supervision.
Extensive experimental results on ShanghaiTech and QNRF datasets demonstrate superior performance over state-of-the-art methods. Code is available at \url{https://github.com/RaccoonDML/CHSNet}.
\end{abstract}



\section{Introduction}
Crowd counting is to count the people from a given image in diverse crowded scenes, which is an active computer vision task with a wide range of promising applications in crowd management, traffic monitoring, surveillance systems, etc. Existing methods can be roughly categorized into detection-based~\cite{rabaud2006counting}, count-based~\cite{chan2009bayesian} and density-map-based~\cite{zhang2016single,li2018csrnet, tian2019padnet, gao2023forget}. The detection-based methods~\cite{rabaud2006counting} require laborious annotations (\ie the bounding boxes) to directly detect all persons in an image while the count-based methods~\cite{chan2009bayesian} only predict the total number of people, suffering from weak supervision.

\begin{figure}
    \centering
    \includegraphics[width=0.9\linewidth]{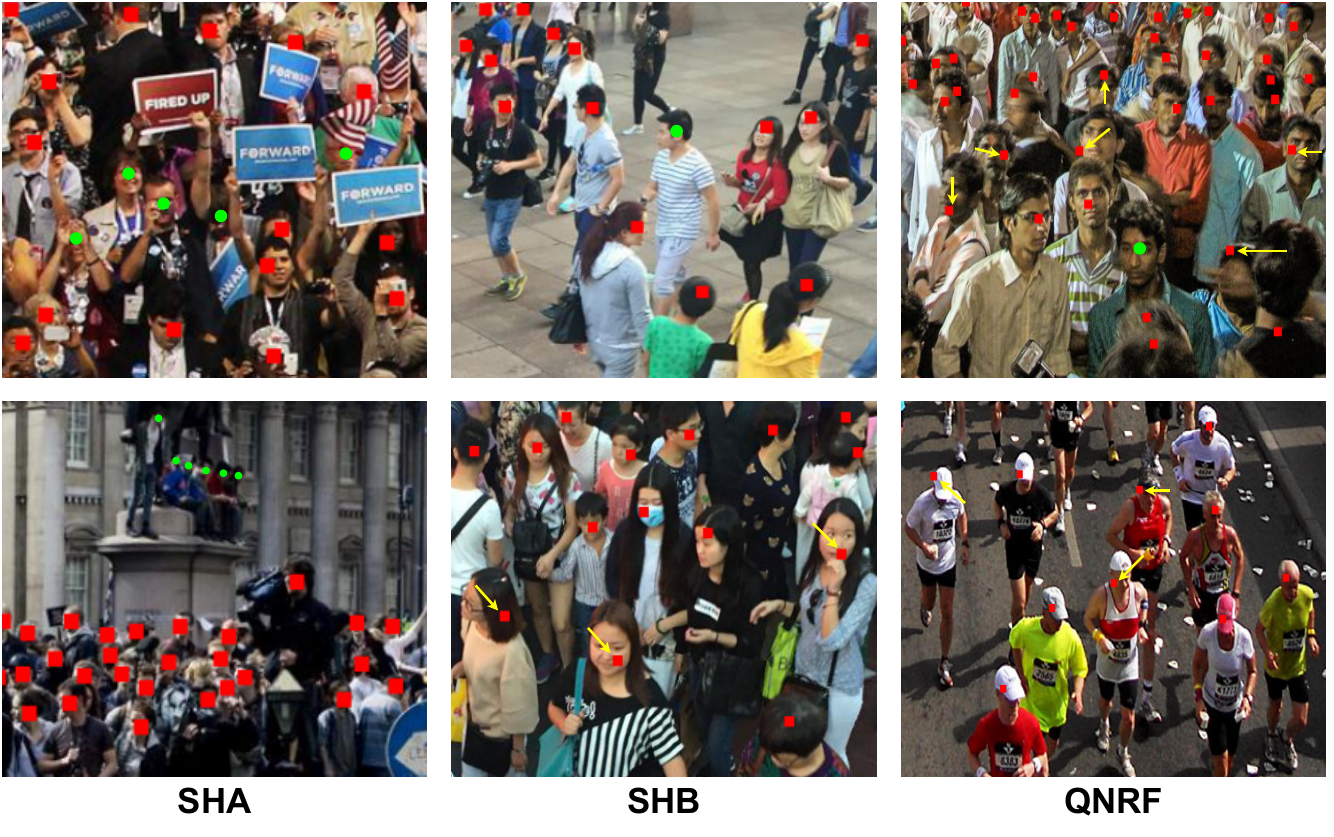}
    \vspace{-10pt}
    \caption{Noisy annotations in commonly-used datasets. Red points, green points, and yellow arrows denote labeled annotations, missing annotations, and location shifts, respectively.
    }
    \label{fig:noisy}
\end{figure}

Unlike the two categories above, 
density map-based approaches~\cite{zhang2016single,li2018csrnet} are proposed to estimate the human densities in images, which can balance the performance and annotation cost. Generating a density map only requires point annotations at the center of each head, whose cost is much less than the detection-based methods. 
In addition, density maps can provide more fine-grained pixel-level supervision compared to 
the count-based methods, which has significantly improved the performance.
However, density map-based methods require accurate point annotations to provide reliable pixel-level supervision, which is usually unrealistic because of the potential noises in the labeling process.
Fig.~\ref{fig:noisy} shows that missing annotations and location shifts commonly exist among widely-used crowd counting datasets, especially in dense scenes and low-resolution conditions. 
Therefore, directly using the pixel-level loss function for optimization may compromise the prediction performance. Furthermore, the counting model may memorize the noisy annotations~\cite{liu2020early}.

\begin{figure*}[t]
    \centering
    \includegraphics[width=0.98\linewidth]{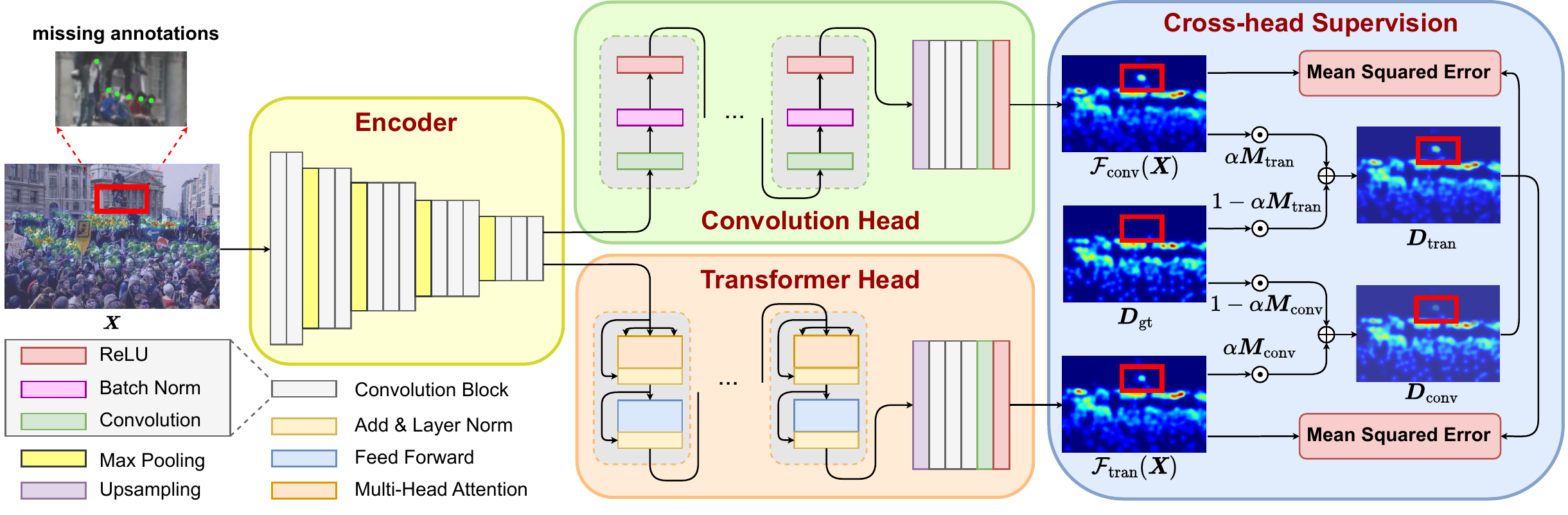}
    \vspace{-10pt}
    \caption{Architecture of \methodname consisting of one shared encoder and two regression heads---one convolution head and one transformer head. At the end of each head, a regression block of the same architecture is attached to produce density map. Red box indicates the area of missing annotations, in which refined supervision is more reliable than ground truth supervision.}
    \label{fig:model}
\end{figure*}

To address this problem, Ma \etal~\cite{ma2019bayesian} designed a Bayesian loss function for instance-level supervision. Cheng \etal~\cite{cheng2019learning} proposed a Maximum Excess over Pixels (MEP) loss function, where the region with the maximum loss value is used for optimization.
Wan \etal~\cite{wan2020modeling} modeled the annotation noise using a random variable with Gaussian distribution. 
Some other methods adopt uncertainty estimation to model the noisy annotations~\cite{oh2020crowd, ranjan2020uncertainty}.
Recently, Lin \etal~\cite{lin2022boosting} proposed instance attention loss to exclude the instance in back-propagation if its deviation is too large.
However, all of these methods above ignore one critical problem: \emph{how to explore useful supervision in noisy areas?}

Our answer is: \emph{the predicted density map itself can provide useful supervision in noisy areas.}
In this paper, we propose to use two regression heads with totally different architecture, \ie one convolution head and one transformer head, to mutually supervise each other in noisy areas.
The resultant model, \methodname, can synergize different types of inductive biases of convolutions and transformers to boost each other's performance for better counting.
However, the quality of the predicted density map is unsatisfactory in the early stage of training and cannot be directly used for supervision. 
To make the training reliable, we develop a progressive cross-head supervision learning strategy, that is, the true supervision density maps should be the weighted combinations of the ground truth and predictions from another head, where the weights are linearly increased as the training process goes on.

The main contributions of this work are summarized as follows:
1) we propose a novel model \methodname with one convolution head and one transformer head to supervise each other in noisy areas;
2) we design a progressive cross-head supervision learning strategy to make the training process more stable; and
3) our \methodname achieves superior performance on several benchmarked datasets.

\section{Methodology}

\subsection{Network Architecture}
Convolutions have strong local modeling ability while transformers~\cite{vaswani2017attention} can effectively capture global context dependencies. We tackle the noisy annotations present in the data and serve the predictions from two heads as each other's supervision in noisy areas. Thus, the negative impact of noisy annotations on training is effectively reduced. The inductive biases of the two heads are fully utilized to complement each other. 

Fig.~\ref{fig:model} presents the proposed \methodname consisting of the following components
: 1) a shared encoder $E$ that extracts the features $E(\mat{X})\in \mathbb{R}^{C\times H\times W}$ from the input image $\mat{X}$; and 2) two regression heads including a convolution head $H_{\mathrm{conv}}$ and a transformer head $H_{\mathrm{tran}}$ to predict the density maps $H_{\mathrm{conv}}\left(E(\mat{X})\right)\in \mathbb{R}^{H\times W}$ and $H_{\mathrm{tran}}\left(E(\mat{X})\right)\in \mathbb{R}^{H\times W}$, where $C$, $H$ and $W$ are the channels, height, and width of the features maps, respectively.
At the end of each head, we adopt an upsampling layer, three convolution blocks, and ReLU activation function to aggregate features and output the density map.
The encoder $E$ adopts VGG16~\cite{simonyan2014very} without the last maxpooling layer and fully-connected layers pre-trained on ImageNet~\cite{deng2009imagenet} for initialization. 

On the one hand, the convolution head $H_{\mathrm{conv}}$ consists of a series of convolution blocks. Each block is stacked by dilated convolution, batch normalization, and ReLU activation layers, which models the local contextual information.
On the other hand, the transformer head $H_{\mathrm{tran}}$ contains several transformer layers, whose inputs are the flattened features of $E(\mat{X})$ along $H$ and $W$, to capture global contextual dependencies of current features.
For the sake of simplification, we omit the intermediate results and use $\mathcal{F}_\mathrm{conv}(\mat{X})$ and $\mathcal{F}_\mathrm{tran}(\mat{X})$ to denote the predictions of two heads.

\subsection{Cross-head Supervision}
Since the two regression heads generate unreliable density maps at the early stage, we propose a progressive cross-head supervision learning strategy to stabilize the training process. The refined supervision of each head is a weighted combination of another prediction and the ground truth density map. The weights of the complementary head are gradually increased as the training proceeds. Formally, the refined supervision of convolution head $\widehat{\mat{D}}_\mathrm{conv}$ is defined as follows:
\begin{equation}
    \widehat{\mat{D}}_\mathrm{conv}=\alpha\mathcal{F}_\mathrm{tran}(\mat{X})+(1-\alpha) \mat{D}_\mathrm{gt},
\end{equation}
where $\alpha$ is the combination coefficient to control the importance of two terms and $\mat{D}_\mathrm{gt}$ is the ground truth density map which is usually mislabeled in some area.

Considering that noisy annotations only account for a small part of all annotations, in our method, only a specific mislabeled area uses the refined supervision, while the other areas should use the original ground truth. 
Typically, the mislabeled examples usually have a large loss~\cite{arazo2019unsupervised,huang2023twin}, as the model would predict the correct labels if it is trained well. Therefore, to select those mislabeled areas, in practice, we first sort the deviation $\mat{\epsilon}=| \mathcal{F}_\mathrm{conv}(\mat{X})-\mat{D}_\mathrm{gt} | \in \mathbb{R}^{W\times H} $ in descending order and obtain the top $\delta\in [0,1]$ value as the mask threshold, denoted as $t_\delta$. In a sense, $\mat{\epsilon}$ describes the discrepancy between ground truth and model prediction. 
Then the selection mask of the convolution head is given by:
\begin{align}
    \mat{M}_\mathrm{conv} = \mathbb{I}(\mat{\epsilon} \geq t_\delta) \in \{0,1\}^{W\times H},
\end{align}
where $\mathbb{I}(\cdot)$ is an indicator function. 

Once we have the selection mask, the final supervision for convolution head can be calculated by:
\begin{align}
    \mat{D}_\mathrm{conv}=\mat{M}_\mathrm{conv}\odot & \widehat{\mat{D}}_\mathrm{conv}+(1-\mat{M}_\mathrm{conv})\odot \mat{D}_\mathrm{gt} \\
    =\alpha \mat{M}_\mathrm{conv} \odot &\mathcal{F}_\mathrm{tran}(\mat{X}) + (1-\alpha \mat{M}_\mathrm{conv}) \odot  \mat{D}_\mathrm{gt},
\end{align}
where $\odot$ represents the element-wise multiplication. Similarly, the final supervision for transformer head $\mat{D}_\mathrm{tran}$ can be calculated using the prediction result by the convolution head. Finally, the overall loss function for optimization is 
\begin{align}
    \mathcal{L}={\| \mathcal{F}_\mathrm{conv}(\mat{X})-\mat{D}_\mathrm{conv} \|}^2_2+
    {\| \mathcal{F}_\mathrm{tran}(\mat{X})-\mat{D}_\mathrm{tran} \|}^2_2.
\end{align}
As a result, the supervision of mislabeled areas is refined from another head.

\subsection{Progressive Learning Strategy}
The predicted density maps of \methodname are unstable in the early stage of training. Therefore, they cannot be directly used for cross-head supervision. To make the early training process stable, we develop a progressive cross-head supervision learning strategy, that is, 
the noise ratio $\delta$ and the combination coefficient $\alpha$ are linearly increased to the preset maximum value as the training process goes on. Formally, the noise ratio and the combination coefficient at the $i$-epoch are calculated as follows:
\begin{align}
    \delta_i=\delta_\mathrm{max}* i/T,\quad
    \alpha_i=\alpha_\mathrm{max}* i/T,
\end{align}
where $\delta_\mathrm{max}$ and $\alpha_\mathrm{max}$ are the predefined maximum noise ratio and maximum combination coefficient, respectively. $T$ denotes the maximum epoch for training.

\section{Experiments}

\subsection{Experiment Setups}

\noindent\textbf{Datasets.}\quad We evaluate our method on three widely-used datasets: ShanghaiTech Part A\&B~\cite{zhang2016single} and UCF-QNRF~\cite{idrees2018composition}. ShanghaiTech Part A\&B contains 482 images (300/182 for training/validation) and 716 images (316/400 for training/validation), respectively. UCF-QNRF includes 1535 high-resolution images (1201/334 for training/validation).
This setting covers from sparse scenes to dense scenes and from small dataset to large dataset.

\noindent \textbf{Evaluation metrics.}\quad As \methodname has two predicted density maps, we use their averaged density map as the final result for evaluation.
Mean Absolute Error (MAE) and Mean Squared Error (MSE) are adopted for evaluation. 

\begin{table}[h]
\centering
\vspace{-10pt}
\caption{Comparisons with the state-of-the-arts methods on SHA, SHB, and QNRF.
}
\scalebox{0.75}{
    \begin{tabular}{rccc}
    \shline
    \multirow{2}{*}{\textbf{Methods}} & \textbf{SHA} & \textbf{SHB} & \textbf{QNRF} \\
    & MAE / MSE & MAE / MSE & MAE / MSE \\
    \hline
    CSRNet~\cite{li2018csrnet}  & 68.2 / 115.0 & 10.6 / 16.0 & - / - \\
    SANet~\cite{cao2018scale}  & 67.0 / 104.5 & 8.4 / 13.6 & - / - \\
    TEDnet~\cite{jiang2019crowd} & 64.2 / 109.1 & 8.2 / 12.8 & 113.0 / 188.0 \\
    BL~\cite{ma2019bayesian}      & 62.8 / 101.8 & 7.7 / 12.7  & 88.7 / 154.8 \\
    DM-Count~\cite{wang2020distribution} & 59.7 / \underline{95.7} & 7.4 / 11.8 & 85.6 / 148.3\\
    MCC~\cite{zand2022multiscale} & 71.4 / 110.4 & 9.6 / 15.0 & - / - \\
    NoisyCC~\cite{wan2020modeling}     & 61.9 / 99.6  & 7.4 / \underline{11.3}  & 85.8 / 150.6\\
    GL~\cite{wan2021generalized}  & 61.3 / \textbf{95.4} & 7.3 / 11.7 & 84.3 / 147.5 \\
    LibraNet~\cite{liu2020weighing}        & \textbf{55.9} / 97.1  & \underline{7.3} / \textbf{11.3}	& 88.1 / \textbf{143.7} \\
    GauNet(CSRNet)~\cite{cheng2022rethinking}  & 61.2 / 97.8 & 7.6 / 12.7	& \underline{84.2} / 152.4 \\
    \textbf{\methodname~(ours)}     & \underline{59.2} / 97.8  & \textbf{7.1} / 12.1  & \textbf{83.4} / \underline{144.9}  \\
    \shline
    \end{tabular}
    }
\label{tab:comparison}
\end{table}

\begin{figure*}[t]
    \centering
    \includegraphics[width=\linewidth]{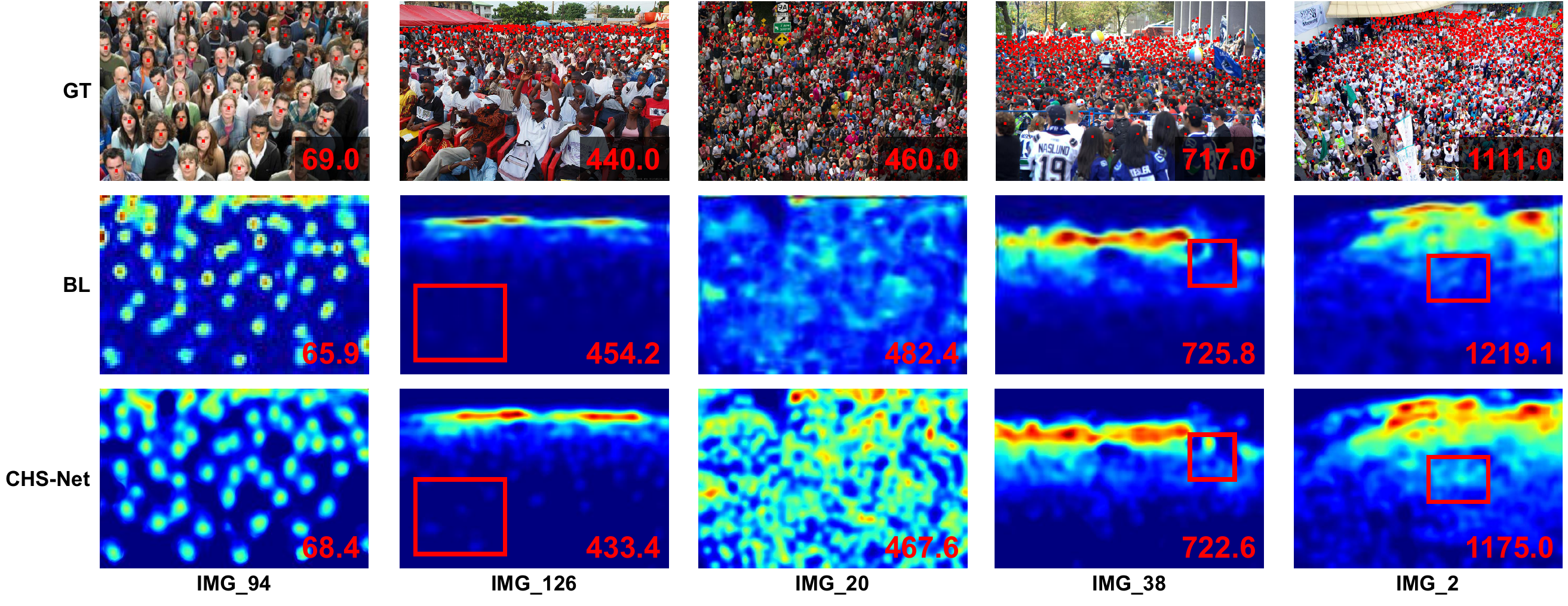}
    \vspace{-20pt}
    \caption{Visualization on SHA. The first row are the input images. The second and third rows are the predicted density maps by BL~\cite{ma2019bayesian} and \methodname, respectively. Red boxes highlight the differences between them.
    }
    \label{fig:visualization}
\end{figure*}

\noindent \textbf{Implementation details.}\quad
We adopt the same data preprocessing in~\cite{ma2019bayesian}. Ground truth density maps are generated using a fixed Gaussian kernel of size 15.
Random scaling, cropping, and horizontal flipping are employed as data augmentation with an image size of $512 \times 512$.
Adam optimizer~\cite{kingma2014adam} was used with an initial learning rate of $4.0\times 10^{-5}$ and weight decay of $1.0\times10^{-5}$. We use the cosine learning rate scheduler with a maximum epoch of 1,000. 
For hyperparameters of cross-head supervision, 
$\delta_\mathrm{max}$ is set to $0.1$ for SHA and $0.05$ for others.
$\alpha_\mathrm{max}$ is set to $0.5$ for QNRF and $1.0$ for others.

\subsection{Comparison with State-of-the-art Methods}
We compare our method with several recent state-of-the-art methods in Table~\ref{tab:comparison}. \methodname consistently achieves the superior counting performance on all benchmark datasets. For SHA and SHB, our method achieves 59.2 and 7.1 in terms of MAE. For QNRF, \methodname improves MAE value of the second best GauNet~\cite{cheng2022rethinking} from 84.2 to 83.4.


The predicted density maps of \methodname are visualized in Fig.~\ref{fig:visualization}. As can be seen, \methodname can predict reliable density maps with high counting accuracy in a wide range of scenes and density levels.


\begin{table}[t]
\centering
\vspace{-10pt}
\caption{Ablation study. `Average' represents using the average density map of two heads for evaluation.}
 \scalebox{0.8}{
		\begin{tabular}{cccrr}  
            \shline
			Model & Cross-head Supervision & Evaluation Head & MAE & MSE\\
			\hline
			Conv. & \xmark & Conv.  & 63.8 & 110.4 \\
			Tran. & \xmark & Tran.  & 62.5 & 104.1 \\
			\hline
			\methodname & \xmark   & Conv. & 61.8 & 107.0 \\ 
			\methodname & \xmark  & Tran. & 60.7 & 103.7 \\
			\methodname & \xmark  & Average & 60.7 & 104.8\\
			\hline
			\methodname & \cmark & Conv. & 59.8 & 100.4 \\ 
			\methodname & \cmark &  Tran. & 60.0 & \textbf{96.7} \\
			\methodname & \cmark &  Average & \textbf{59.2} & 97.8 \\
			\shline
		\end{tabular}
  }
    \label{tab:ablation}
\end{table}

\subsection{Ablation Studies}

In this section, we perform ablation studies on SHA dataset to evaluate the effectiveness of the proposed \methodname.

\noindent\textbf{Ablation study on the models.}\quad
We first evaluate the model with two heads and cross-head supervision. We have the following observations from  Table~\ref{tab:ablation}:
1) The MAE of transformer head is lower than convolution one by 1.3, which demonstrates the superior global modeling ability of the transformer. 
2) Although without cross-head supervision, \methodname has achieved significant improvements.
3) With the progressive cross-head supervision learning strategy, the best performance is achieved with MAE of 59.2. 
4) We evaluate the performance of every single head in \methodname. The average of two heads outperforms any single head in terms of MAE, which further illustrate the advantage of \methodname.

\noindent\textbf{Ablation study of two heads on easy/hard samples.}\quad Our method is not only an ensemble model with different heads but also contains a self-supervision mechanism implicitly, in which the two heads provide pseudo-labels for each other in the mislabeled area.
Based on this idea, we would highlight that the convolution head and transformer head have different learning capabilities.
Here, we simply split the hard/easy samples according to the number of humans in an image, \emph{i.e.} the 50\% samples with the most humans are hard ones, and we showcase their performance to validate our idea in Table~\ref{tab:easy-hard}. 
It is found that the convolution head is better at learning easy samples and the transformer head plays the opposite role, respectively. Therefore, the different learning capabilities of these two heads are of great benefit for providing supervision in the mislabeled area for each other.

\begin{table}[h]
    \centering
    \vspace{-10pt}
    \caption{Ablation study of two heads on easy/hard samples.}
    \scalebox{0.8}{
    \begin{tabular}{c|c|c}
    \shline
    MAE / MSE & Easy samples & Hard samples  \\
              \hline
    Conv-head & \textbf{55.7} / \textbf{70.5}  & 216.5 / 299.6 \\
    \hline
    Tran-head & 65.2 / 83.4  & \textbf{159.1} / \textbf{204.3}\\
    \shline
    \end{tabular}
    }
    \label{tab:easy-hard}
\end{table}

\noindent\textbf{Effect of maximum noise ratio.}\quad The maximum noise ratio $\delta_\mathrm{max}$ is an important hyperparameter for \methodname. In fact, the maximum noise ratio is a kind of prior knowledge of a specific dataset. We set several values of maximum noise ratio to investigate its effect on model performance. In Table~\ref{tab:noise}, the best performance is obtained when $\delta_\mathrm{max}=0.1$, which means there are nearly 10\% noisy annotations in SHA.

\begin{table}[h]
\centering
    \vspace{-10pt}
    \caption{Effect of maximum noise ratio $\delta_\mathrm{max}$.}
    \scalebox{0.8}{
	\begin{tabular}{crrrrrr}
        \shline
		$\delta_\mathrm{max}$ & 0 & 0.01 & 0.05 & 0.10 & 0.15  & 0.30 \\
		\hline
		MAE & 60.7 & 60.8 & 61.1 & \textbf{59.2} & 60.5 & 61.0 \\
		MSE & 104.8 & 105.7 & 99.8 & \textbf{97.8} & 102.4 & 105.4 \\
		\shline
	\end{tabular}
 }
	\label{tab:noise}
\end{table}

\vspace{-10pt}
\section{Conclusion}
\vspace{-5pt}
Noisy annotations are common in crowd counting datasets. To alleviate the negative impact of noisy annotations, we propose \methodname, a network with a convolution head and a transformer head to mutually supervise each other in noisy areas. In addition, we develop a progressive cross-head supervision learning strategy to stabilize training process and provide more reliable supervision. Experimental results show the superior performance of our proposed approach.
For future work, we will explore more noise robust loss functions to further utilize the ability of \methodname. Besides, we may consider to enhance the model sustainable learning ability like~\cite{gao2023forget} so that the noise ratio of training samples from different domains can be adaptively adjusted.

\clearpage
\bibliographystyle{IEEEbib}
\bibliography{refs}

\end{document}